\documentclass{article}
\usepackage{iclr2016_workshop,times}

\usepackage{graphicx}
\usepackage{hyperref}
\usepackage{url}
\usepackage{wrapfig}

\title{On-the-fly Network Pruning\\ for Object Detection}

\author{Marc Masana, Joost van de Weijer \& Andrew D. Bagdanov \\
Computer Vision Centre\\
Universitat Aut\`{o}noma de Barcelona\\
Barcelona, 08193, Spain\\
\texttt{\{mmasana,joost,bagdanov\}@cvc.uab.cat}}

\newcommand{\minisection}[1]{\vspace{0.04in} \noindent {\bf #1}\ \ }

\begin{document}
\maketitle
\begin{abstract}
  Object detection with deep neural networks is often performed by
  passing a few thousand candidate bounding boxes through a deep
  neural network for each image. These bounding boxes are highly
  correlated since they originate from the same image. In this paper
  we investigate how to exploit feature occurrence at the image scale
  to prune the neural network which is subsequently applied to all
  bounding boxes. We show that removing units which have near-zero
  activation in the image allows us to significantly reduce the number
  of parameters in the network. Results on the PASCAL 2007 Object
  Detection Challenge demonstrate that up to 40\% of units in some
  fully-connected layers can be entirely eliminated with little change
  in the detection result.
\end{abstract}
\section{Introduction}
Deep neural networks are often trained for recognition problems over
very many labels. This is partially to ensure wide applicability of
the network and partially because networks are known to benefit
from multi-label data (additional training examples from one class can
increase performance of another class because they share features
among several layers). At testing time, however, one might want to
apply the neural network to a collection of examples which are
highly correlated. They only contain a limited subset of the
original labels and consequently will result in sparse node
activations in the network. In these cases, application of the full
neural network to the whole collection results in a considerable
amount of wasted computation. In this paper we describe a method for
pruning of neural networks based on analysis of internal unit
activations with the objective of constructing more efficient
networks.

In computer vision many problems have the structure described
above. We briefly mention two here. Imagine you want to classify the
semantic content in each frame (an example) of a video (the
collection). A fast assessment of the video might reveal that it is an
indoor birthday party. This knowledge might exclude many of the nodes
in the neural network -- those which correspond to 'snow', 'leopards',
and 'rivers', for example, will be unlikely to be needed in any of the
thousands of frames in this video. Another example is object
detection, where we extract thousands of bounding boxes (examples)
from a single image (the collection) with the aim of locating all
semantic objects in the image. Given an assessment of the image, we
have knowledge of the node activations for the entire collection, and
based on this we can propose a smaller network which is subsequently
applied to the thousands of bounding boxes. We will here only consider the latter
example in more detail.

Reducing the size and complexity of neural networks (or network
\emph{compression}) enjoys a long history in the learning community.
The authors of~\cite{bucila2006model} train a simpler neural network
to mimic the output of a complex one, and in~\cite{ba2014deep} the
authors compress deep and wide (i.e. with many feature maps) networks
to shallow but wider ones. The technique of Knowledge Distillation was
introduced in~\cite{hinton2015distilling} as a model compression
framework. The framework compresses an ensemble of deep networks
(teacher) into a student network of similar depth. More recently, the
FitNets approach leverages the Knowledge Distillation framework to
exploit depth and train student networks that are \emph{thin} but
remain \emph{deep}(~\cite{AdrianaFitNets:2014}). Another network
compression strategy was proposed in~\cite{girshick2015fast,xue2013restructuring} that uses
singular value decomposition to reduce the rank of weight matrices in
fully connected layers in order to improve efficiency.

In this paper we are not interested in mimicking the operation of a
deep neural network over all examples and all classes (as in the
student-teacher compression paradigm common in the
literature). Rather, our approach is to make a quick assessment of
image content and then, based on analysis of unit activation on entire
image, to modify the network to use only those units likely to
contribute to correct classification of labels of interest when
applied to each candidate bounding box.

\begin{wrapfigure}{r}{0.5\textwidth}
  \begin{center}
    \includegraphics[width=0.45\textwidth]{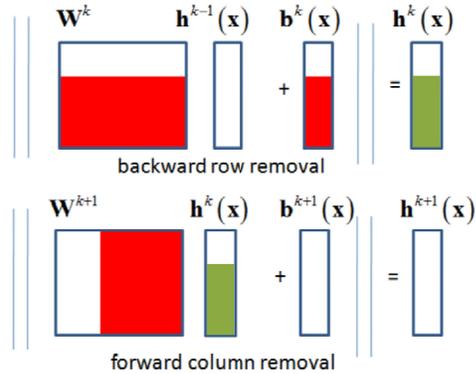}
  \end{center}
  \caption{Example of backward and forward unit pruning. We use $\left\| . \right\|$ to indicate the $\mathrm{relu}\left(.\right)$ activation function. Based on knowledge that some unit activations ${\bf{h}}^k \left( {\bf{x}} \right)$ are zero (indicated in green), we can reduce the parameters of ${\bf{W}}^k$, ${\bf{W}}^{k + 1}$ and ${\bf{b}}^k$ (indicated in red).}\label{fig:pruning}
\end{wrapfigure}
\section{Forward and backward unit pruning for object detection}
Consider the original neural network $f\left( {x;\theta } \right)$,
where $\theta$ are the network parameters. We wish to compute a
network defined by parameters $\theta^*$ for which:
\begin{equation}
f\left( {x;\theta ^* } \right) \approx f\left( {x;\theta } \right)\forall x \in C
\end{equation}
where $\left| {\theta ^* } \right| < \left| \theta \right|$ (i.e.  the
number of parameters in $\theta^*$ is considerably lower than in the
original network). In the case of object detection we will use the unit
activations of the entire image to prune the network which will be
applied to all the bounding box proposals. This is based on the
observation that for some layers, nodes with zero activations on the
whole image cannot have nonzero activation on any bounding box in the image.

The hidden layer activation of a fully connected layer $k$ can be
written as:
\begin{equation}
{\bf{h}}^k \left( {\bf{x}} \right) = \mathrm{relu}( {{\bf{b}}^k  + {\bf{W}}^k {\bf{h}}^{k - 1} \left( {\bf{x}} \right)} )
\end{equation}
where ${\bf{b}}^k$ and ${\bf{W}}$ are the biases and weights of the
$k$-th layer, and $\mathrm{relu}(\cdot)$ indicates the rectified
linear activation function.  We first consider how knowledge of the
absence of node activations in the image can be
translated into a network with fewer parameters. We consider two
cases: backward and forward unit pruning, as illustrated in Fig.~\ref{fig:pruning}.

\minisection{Backward unit pruning:} Without loss of generality, we
order the activations in layer $h^k$ so that the $q$ non-active, zero
nodes are at the end of vector $h^k$. Then we can write:
\begin{equation}
\left[ {{\bf{h}}^k \left( {\bf{x}} \right)_{1:\left( {n - q} \right)} ;{\bf{0}}_{q,1} } \right] = \mathrm{relu}\left( {\left[ {{\bf{W}}^k _{1:\left( {n - q} \right),1:m} ;{\bf{0}}_{q,m} } \right]{\bf{h}}^{k - 1} \left( {\bf{x}} \right) + \left[ {{\bf{b}}^k _{_{1:\left( {n - q} \right)} } ;{\bf{0}}_{q,1} } \right]} \right)
\label{eq:backward}
\end{equation}
where we use ${\bf{0}}_{m,n}$ to indicate the zero-matrix of
dimension $m$ by $n$, and subscripts are used to indicate a selection
of indices from the original vector or matrix. We use
$\left[ {.,.} \right]$ for horizontal and $\left[ {.;.} \right]$ for
vertical concatenation (following Matlab
convention). Eq.~\ref{eq:backward} shows that backward unit pruning
allows us to remove from ${\bf{W}}^k$ and ${\bf{b}}^k$ an equal amount
of rows as there are zeros in ${\bf{h}}^k$ -- \emph{without changing 
the output of the network.}

\minisection{Forward unit pruning:} Here we look how the zeros in the
activation ${\bf{h}}^k$ can be exploited to remove parameters from the
following layer. The activation in layer $k+1$ can be written:
\begin{equation}
{\bf{h}}^{k + 1} \left( {\bf{x}} \right) = \mathrm{relu}\left( {\left[ {{\bf{W}}^{k + 1} _{1:p,1:\left( {n - q} \right)} ,{\bf{0}}_{p,q} } \right]\left[ {{\bf{h}}^k \left( {\bf{x}} \right)_{1:\left( {p - q} \right)} ;{\bf{0}}_{q,1} } \right] + {\bf{b}}^{k + 1} } \right)
\end{equation}
In this case, the zeros in ${\bf{h}}^k$ result in the removal of
columns from ${\bf{W}}^{k + 1}$. These can be removed without changing
the output of the network.

In practice there might only be a few zero activation in the image
 and therefore we consider all node activations which are below a certain threshold to be zero\footnote{In case the activation function is not the ReLU one should consider the absolute value of
  the activation function to be smaller than a threshold.}. This
allows us to further increase the parameter reduction of the network $f\left( {x;\theta ^* } \right)$
 but at the cost of slight deviations from the original
network $f\left( {x;\theta } \right)$. We also note that although notations are about fully-connected layers for simplicity, our proposal would also be applicable to convolutional layers too.
\section{Results and Conclusions}
We evaluate our proposed methods on the VOC PASCAL 2007
dataset~(\cite{everingham2010pascal}) with the fast R-CNN
framework by~\cite{girshick2015fast}. The VOC 2007 has a total of 24,640
annotated objects for training, with an average of 2.5 objects per
image, and in the test set an average of 2.4 objects per image. The
Fast R-CNN framework is fit for our purposes since it first passes the
image through all the convolutional layers to later use the extracted
feature maps with the corresponding bounding boxes which we want to
evaluate (usually 1,000+ boxes). The network used a modification of
the VGG16 network~(\cite{simonyan2014very}).

\minisection{Forward pruning.} Our first experiment uses forward unit
pruning on the \textit{pool5} layer of the VGG16 network to reduce the
number of parameters of the \textit{fc6} layer. This is the layer with
highest percentage of parameters (38.7\% parameters in the
network). The \textit{pool5} layer has $512\times 7\times 7$ outputs,
where the first dimension represents the feature maps, and the second
and third dimensions are spatial dimensions (smaller than the original
image size because of the resizing at each pooling layer). In order to
decide which activations to prune, we first pass the whole image
through the network and observe the activations at each unit in
\textit{pool5}. We sum over the spatial dimensions and apply a
threshold to select units to prune from the network before applying it
to all bounding boxes.
\begin{wrapfigure}{l}{0.48\textwidth}
  \begin{center}
    \includegraphics[width=0.45\textwidth]{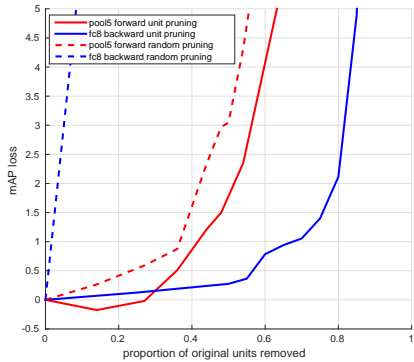}
  \end{center}
  \caption{Performance loss as a function of parameter reduction.}\label{fig:results}
\end{wrapfigure}
Results show an initial minor improvement in the performance of the
framework when removing parameters (see Fig.~\ref{fig:results}). The
lack of propagation through the network of very low value activations
could be the cause of the small difference in performance. Then, for
reductions of 25-40\% of the parameters on layer \textit{fc6}, we
obtain a mAP loss of less than 1. From that point on, further removal
of parameters leads to higher loss. This happens because the
activations removed start to be too relevant for the network's
discriminative
power.

\minisection{Backward pruning.} The second experiment applies backward
unit pruning to the \textit{fc8} layer to reduce the number of
parameters from the weight and bias matrices used to compute the
network outputs. In this case, we use an image classifier (VGG16 deep
features based) to decide which classes (activations) would be more
likely to appear in the original image. Based on that classification,
we adopt a top-$N$ strategy where we keep the $N$ classes with higher
probability from the image classifier and remove the rest. This
reduction affects the weight and bias matrices of the \textit{fc8},
which would no longer propagate into the following layers (the softmax
in this case).  In this case, results keeping 6 or more classes
(reductions of 0-70\%) show a mAP loss of less than 1. However,
performance starts dropping after because of images having more
classes present than classes kept. It should be noted that only a
small percentage of the total parameters of the network are in
\textit{fc8}. However, when considering object detection with
thousands of classes, the relevance of this layer is comparable to
\textit{fc6}.

\minisection{Conclusions.} We have presented a method to prune units
in neural networks for object detection through analysis of unit
activation on the entire image. We show that for some layers up to
40\% of the parameters can be removed with minimal impact on
performance. We are interested in combining our method with other
parameter reduction methods such as~\cite{xue2013restructuring}. Also
applying our method to other types of layers (e.g. convolutional) and
evaluating on datasets with very many labels are promising research
directions. In addition, we are interested in applying our method to semantic segmentation where, similarly as in our problem, a redundant network is applied to every pixel.

\subsubsection*{Acknowledgments}
This work is funded by the Projects TIN2013-41751-P of the Spanish Ministry of Science, the Catalan project 2014 SGR 221 and the CHIST ERA project PCIN-2015-226. We gratefully acknowledge the support of NVIDIA.

\bibliography{iclr2016_workshop}
\bibliographystyle{iclr2016_workshop}

\end{document}